\documentclass[10pt,twocolumn,letterpaper]{article}

\usepackage{cvpr}              % To produce the CAMERA-READY version
\usepackage{microtype}
\usepackage{adjustbox}
\usepackage{booktabs}
\usepackage{graphicx}
\usepackage{amsmath}
\usepackage{amssymb}
\usepackage{colortbl}
\usepackage{dblfloatfix}

\makeatletter
\g@addto@macro\normalsize{%
  \setlength\abovedisplayskip{5pt plus 2pt minus 2pt}%
  \setlength\belowdisplayskip{5pt plus 2pt minus 2pt}%
  \setlength\abovedisplayshortskip{2pt plus 1pt}%
  \setlength\belowdisplayshortskip{3pt plus 1pt minus 1pt}%
}
\makeatother

\makeatletter
\renewcommand\paragraph{\@startsection{paragraph}{4}{\z@}%
  {4pt plus 1pt minus 1pt}{-1em}{\normalfont\normalsize\bfseries}}
\makeatother

\definecolor{cvprblue}{rgb}{0.21,0.49,0.74}
\usepackage[pagebackref,breaklinks,colorlinks,allcolors=cvprblue]{hyperref}

\def\confName{CVPR}
\def\confYear{2027}

\title{ORGANIC-3D: Grounding 3D CT Vision Language Models in Organ-Level Image–Text Coupling}

\author{
Zelong Liu \quad Hamza Ahmed \quad Wynn Wingyi Lee \quad Mickael Torjman \quad Alexander Dimitrakakis\\
Jiahua Lu \quad Timothy Deyer \quad Bachir Taouli \quad Zahi A. Fayad \quad Xueyan Mei\\[4pt]
Icahn School of Medicine at Mount Sinai, New York, NY, USA\\
{\tt\small zelong.liu@icahn.mssm.edu, hamza.ahmed@icahn.mssm.edu, xueyan.mei@mssm.edu}
}
\begin{document}
\maketitle
\begin{abstract}
Vision language models (VLMs) have achieved strong performance in 3D computed tomography (CT) classification, but high accuracy does not reflect whether their predictions rely on the correct anatomical evidence. We introduce \textbf{ORGANIC-3D}, a framework for evaluating and improving organ-level image-text alignment through controlled anatomical interventions. Our benchmark assesses 12 vision encoder checkpoints and 7 generative VLMs using matched-patient comparisons, organ erasure, lesion replacement, and laterality perturbations, revealing substantial alignment failures even among top-performing models. To learn organ-level image-text correspondence, \textbf{ORGANIC-3D} aligns each finding sentence with the anatomy-aware image features and trains on edited scans to anchor this alignment in the organ's image evidence. We further extend this framework to generative VLMs, promoting evidence-dependent question answering and report generation. Trained exclusively on public data, \textbf{ORGANIC-3D} achieves state-of-the-art classification AUROC on CT-RATE, Merlin, and the held-out RAD-ChestCT dataset and reaches 77\% laterality accuracy, compared with at most 66\% for prior models. Its generative VLM matches or exceeds all evaluated publicly released VLMs on visual question answering, achieves the highest report accuracy, and retracts 96–99\% of target-organ findings after organ removal, versus a maximum of 69\% for other models. Together, these results establish \textbf{ORGANIC-3D} as a framework for testing whether 3D CT vision language models ground their predictions in anatomical evidence and for improving organ-level alignment alongside predictive accuracy. Github repo: \url{https://github.com/lzl199704/organic3d/tree/main}.
\end{abstract}
    
\section{Introduction}
\label{sec:intro}

\begin{figure}
\centering
\includegraphics[width=\linewidth]{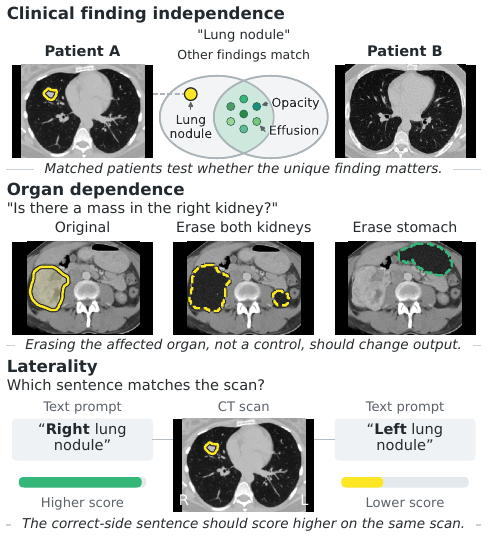}
\caption{\textbf{Alignment tests with expected responses.}
Each test checks whether predictions follow anatomical evidence.
\textbf{Clinical finding independence:} among patients matched on all other report labels, the finding should score higher when present.
\textbf{Organ dependence:} erasing the finding's organ should lower its score toward its prior; erasing a control organ should leave it unchanged.
\textbf{Laterality:} on the same scan, the correct-side sentence should score higher than its side-swapped version.}
\label{fig:tasks}
\end{figure}

Vision language models (VLMs) show promise in 3D computed tomography (CT), from disease classification to visual question answering and report generation~\cite{hamamci2024ctrate,blankemeier2024merlin,bai2024m3d,nvreasonct2026}.
These include contrastive encoders that align volumes with text~\cite{hamamci2024ctrate,blankemeier2024merlin,colipri,spectre,jolia2026} and generative models that produce diagnostic answers or reports~\cite{wu2023radfm,bai2024m3d,medgemma15,nvreasonct2026}.
However, diagnostic accuracy and text quality do not establish whether predictions depend on the relevant anatomical evidence.

VLMs may instead exploit language priors, co-occurring findings, or acquisition-related cues~\cite{geirhos2020shortcut,degrave2021covid}.
Studies in 3D CT report challenges with small lesions~\cite{chevli2026bigbright}, reports influenced by language templates~\cite{bumgardner2026grafting,mayelasserre2026templates}, and difficulty with spatial and left--right reasoning~\cite{monon2026ctspatialvqa}.
In some settings, removing the image can even improve question-answering performance~\cite{liu2026oncovqa}.
These findings motivate evaluating model responses to changes in visual evidence.

Controlled image-text interventions test whether predictions respond appropriately when image or text evidence is altered~\cite{lotfinia2026image,medvil2026}.
Unlike conventional spatial grounding, which directly localizes findings~\cite{rexgroundingct}, controlled edits test whether its prediction changes when the image evidence for the finding is removed or altered.
CT's volumetric structure and anatomical segmentation~\cite{wasserthal2023totalsegmentator} enable organ-level interventions and laterality tests, complemented by comparisons between patients matched on other finding labels (Fig.~\ref{fig:tasks}).

We introduce a benchmark spanning classification, clinical finding independence, organ dependence, and laterality.
We evaluate eleven publicly available vision language encoders across five CT datasets and seven public generative VLMs on CT-RATE~\cite{hamamci2024ctrate}, Merlin~\cite{blankemeier2024merlin}, ReXGroundingCT~\cite{rexgroundingct}, Rad-ChestCT~\cite{draelos2021radchest}, and AbdomenAtlas 3.0~\cite{bassi2025radgpt}.
Our analysis reveals discrepancies between diagnostic performance and image-text alignment, including limited sensitivity to organ removal and frequent preference for the wrong side of a finding.
To address these discrepancies, we develop ORGANIC-3D (ORGAN-level Image-text Coupling in 3D), which combines organ-indexed visual tokens with alignment objectives for target organ removal and spatial transformations.
We extend these principles to generative VLM training to encourage patient-specific answers that depend on relevant anatomical evidence. Our main contributions are:
\begin{itemize}
    \item \textbf{A systematic benchmark for controlled-edit image-text alignment in 3D CT.}
    We introduce four tasks, with organ and lesion erasure as controlled edits, to evaluate publicly available 3D CT models, including eleven vision language encoders and seven VLMs. We show that diagnostic accuracy and image-text alignment can diverge substantially. Publicly encoders reach at most 66\% laterality accuracy, while organ removal produces only limited changes in discrimination performance for ungated encoders.
    \item \textbf{ORGANIC-3D, an anatomy-aware encoder with image-text coupling learning.}
    We introduce anatomy queries, organ routing and text-conditioned pooling to align the image evidence with correct description, and the ORGANIC losses make the model's prediction depend on the organ that holds the finding. Trained on public data only, ORGANIC-3D achieved the highest classification AUROC of all public encoders on every test set and the strongest image-text alignment. Ablations show that the anatomy queries and routing give laterality and the ORGANIC losses double organ dependence without loss of accuracy.
    \item \textbf{A generative 3D CT VLM whose answers and reports depend on the organ.}
    Built on the frozen ORGANIC-3D encoder with a evidence-dependent loss, an answer-attention loss and answer-first reports, the ORGANIC-3D VLM achieves competitive diagnostic performance and report generation.
    We will release the evaluation framework, ORGANIC-3D code and model weights.
\end{itemize}

\section{Related Work}
\label{sec:related}

%-------------------------------------------------------------------------
\subsection{3D CT vision-language models}
\label{sec:related_vlm}
\begin{figure*}[t]
\centering
\includegraphics[width=\linewidth]{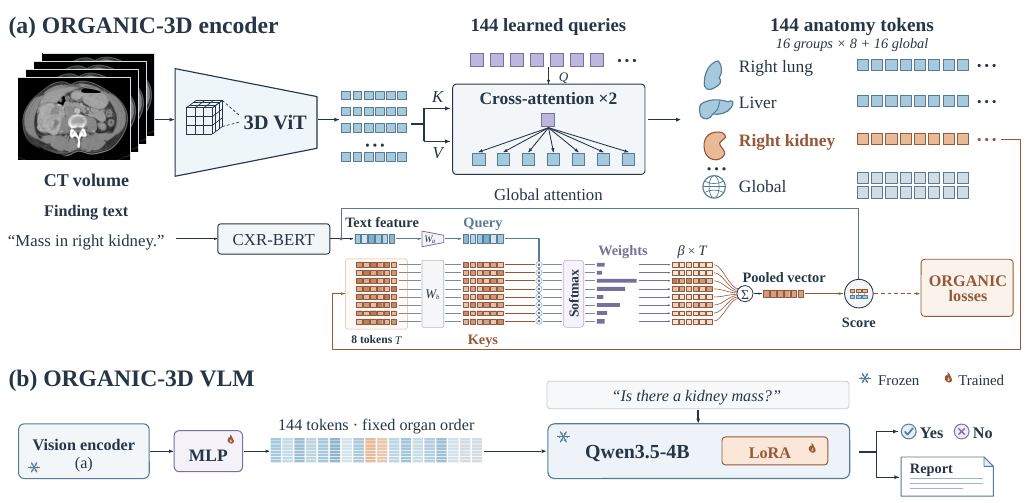}
\caption{\textbf{ORGANIC-3D overview.}
(a)~Cross-attention forms 144 organ and global tokens from 3D ViT features. CXR-BERT text features condition pooling over the named organs' tokens for image--text scoring.
(b)~The frozen vision encoder supplies all anatomy tokens in a fixed organ order to Qwen3.5-4B through an MLP. Only the MLP and LoRA adapters are trained.}
\label{fig:overview}
\end{figure*}
Public 3D CT datasets with paired reports~\cite{hamamci2024ctrate,blankemeier2024merlin} support two model families.
Contrastive encoders align volumes and reports for zero-shot finding classification.
Following CT-CLIP~\cite{hamamci2024ctrate} and Merlin~\cite{blankemeier2024merlin}, encoders scale pretraining~\cite{spectre,hlip} or use finding-specific text objectives~\cite{colipri,itemizedclip}.
Organ masks enter at different stages: fVLM and RADAR pool features within test-time masks~\cite{shui2025fvlm,radar2026}, Jolia learns one mask-free query per anatomical concept~\cite{jolia2026}, and others use masks only during training~\cite{ocpct2026,arcct2026}.
Generative models connect 3D encoders to language models for reporting and VQA, from RadFM and M3D~\cite{wu2023radfm,bai2024m3d} to CT-CHAT~\cite{hamamci2024ctrate}, MedGemma~\cite{medgemma15}, Hulu-Med~\cite{hulumed}, and NV-Reason-CT, whose rewards check listed abnormalities~\cite{nvreasonct2026}.
Evaluation uses zero-shot AUROC for encoders and answer accuracy and report metrics for generative models.
Benchmarks test spatial reasoning, including laterality, and link diagnoses to recognition and measurement~\cite{monon2026ctspatialvqa,chen2026deeptumorvqa}, but use unmodified volumes.
These evaluations cannot distinguish anatomical evidence dependence from co-occurrence or language shortcuts.

%-------------------------------------------------------------------------
\subsection{Image-text alignment}
\label{sec:related_alignment}

Report-supervised pretraining aligns whole images and reports~\cite{zhang2022convirt,boecking2022biovil}, report words and image regions~\cite{huang2021gloria}, and, in 3D CT, individual findings, report sections, and organs~\cite{ctglip,itemizedclip,jolia2026,okact2026}.
Alignment is usually inferred from classification accuracy, sentence grounding to radiologist-segmented findings in ReXGroundingCT~\cite{rexgroundingct}, or attention maps~\cite{jolia2026,ocpct2026}.
Fixing clinical context also changes chest radiograph performance~\cite{wang2026clinicalcontext}.
None establishes dependence on the aligned evidence.

Counterfactual audits edit 2D images or individual slices of volumetric data.
Lotfinia \etal~\cite{lotfinia2026image} occlude annotated findings in chest radiographs against equal-size control boxes and swap images between same-label patients.
SHOVIR perturbs boxes around target or co-occurring findings during report generation and finds that the best generators are not the most image-dependent~\cite{ruffini2026shovir}.
MedVIGIL masks regions of interest or mirrors images~\cite{medvil2026}, and counterfactual grounding exposes hallucinated findings~\cite{song2026cfground}.
Volumetric audits flip and rotate CT slices~\cite{wolf2025otherleft} or remove lesion slices from brain MRI, after which VLMs retain their diagnosis~\cite{khoramfar2026robustness}.
No existing audit erases whole organs in 3D against organ-level controls or applies the same tests to contrastive encoders and generative VLMs.

%-------------------------------------------------------------------------
\subsection{Interventional learning in VLMs}
\label{sec:related_interventional}

Interventional learning trains models to change predictions after evidence removal but preserve them under irrelevant edits.
In VQA, similar images with different answers~\cite{goyal2017vqav2} and object-masked images~\cite{chen2020css} discourage question-only answers.
Generated counterfactuals teach classifiers to reject labels after object removal~\cite{chang2021counterfactual}; interchange intervention training constrains responses to internal activation swaps~\cite{geiger2022iit}.
Generative VLMs use preference optimization with original and corrupted or edited images~\cite{wang2024mdpo,xie2024vdpo,wu2025svco}.
Medical variants use lesion-noised or lesion-masked images~\cite{mmedpo2025,comedpo}, hard-negative and swapped images~\cite{coral2026}, and pathology slides with opposite findings, unmatched on other findings~\cite{zhang2026pairdpo}.

Retaining labels after masking encourages inference from remaining context~\cite{covert2021removing}, while models may recognize edits rather than findings~\cite{xing2026radcf}.
Most objectives retain labels or treat edited examples as negative or less preferred.
Erasing a finding's organ removes evidence without establishing organ health. ORGANIC-3D therefore compares responses to target and control organ removal to learn evidence dependence.
\section{Methods}
\label{sec:method}

% Table 1: encoders, classification and alignment side by side. Load with \input{sec/tables/tab_encoders}.
% Needs \usepackage{colortbl} in preamble.tex.
\begin{table*}[!t]
\centering
\footnotesize
\setlength{\tabcolsep}{3pt}
\renewcommand{\arraystretch}{0.9}
\setlength{\aboverulesep}{1.5pt}\setlength{\belowrulesep}{1.5pt}
\setlength{\abovecaptionskip}{2pt}
\definecolor{oursbg}{RGB}{226,234,248}%  ORGANIC-3D rows (same light blue as Fig. 2)
\definecolor{fail}{RGB}{250,222,218}%    at or below chance on an alignment test
\definecolor{grpA}{RGB}{242,242,242}%    no organ masks
\definecolor{grpB}{RGB}{253,236,218}%    organ masks at test time
\definecolor{grpC}{RGB}{226,242,226}%    fine-tuned on the same public data
\newcommand{\cA}{\rowcolor{grpA}}%
\newcommand{\cB}{\rowcolor{grpB}}%
\newcommand{\cC}{\rowcolor{grpC}}%
\newcommand{\cD}{\rowcolor{oursbg}}%
\newcommand{\cE}{\rowcolor{oursbg!45}}%
\newcommand{\fl}[1]{\cellcolor{fail}#1}%
\caption{\textbf{Encoders evaluation} (all values in \%).
Classification: AUROC, mean over each dataset's findings (Secs.~\ref{sec:metrics} and~\ref{sec:datasets}) on CTR (CT-RATE), RAD (RAD-ChestCT), MER (Merlin), MER-r (the released Merlin labels) and ReX (ReXGroundingCT); AA3: zero-shot tumor detection on AbdomenAtlas~3.0.
Alignment (Sec.~\ref{sec:metrics}): matched pair AUROC (MP), organ erasure $\Delta$AUROC (OE), laterality accuracy (Side) and CT-RATE lung lobe erasure hit rate (LL). Lesion Erasure: lesion erasure hit rate (LE).
COLIPRI-FT: the COLIPRI backbone fine-tuned on the same public data, without ORGANIC losses. Bold: best; underlined: second best.}
\label{tab:encoders}
\begin{tabular}{@{}l cccccc ccccccccccc@{}}
\toprule
 & \multicolumn{6}{c}{Classification} & \multicolumn{9}{c}{Alignment} & \multicolumn{2}{c}{Lesion Erasure} \\
\cmidrule(lr){2-7} \cmidrule(lr){8-16} \cmidrule(lr){17-18}
 & \multicolumn{5}{c}{MLC} & Tumor & \multicolumn{4}{c}{CTR} & \multicolumn{2}{c}{RAD} & \multicolumn{3}{c}{MER} & \multicolumn{2}{c}{LE} \\
\cmidrule(lr){2-6} \cmidrule(lr){7-7} \cmidrule(lr){8-11} \cmidrule(lr){12-13} \cmidrule(lr){14-16} \cmidrule(lr){17-18}
Model & CTR & RAD & MER & MER-r & ReX & AA3 & MP & OE & Side & LL & MP & OE & MP & OE & Side & ReX & AA3 \\
\midrule
\cA CT-CLIP & 73.2 & 62.5 & 57.1 & 59.8 & 62.9 & 59.6 & 65.5 & 1.9 & 55.2 & 8.5 & 59.4 & 3.3 & 58.4 & 0.4 & 50.9 & 55.9 & 46.9 \\
\cA CT-VocabFine & 76.0 & 65.3 & 57.0 & 59.1 & 64.6 & 60.8 & 65.9 & 5.0 & 49.5 & 19.5 & 61.6 & 3.9 & 58.2 & 0.3 & 50.6 & 51.1 & 44.8 \\
\cA HLIP & 77.1 & 70.4 & 58.0 & 62.3 & 67.0 & 44.0 & 71.3 & 11.0 & 60.2 & 12.6 & 68.9 & 12.0 & 59.7 & $-$0.5 & 52.1 & 66.6 & 56.6 \\
\cA ItemizedCLIP & 82.7 & 74.4 & 58.6 & 64.1 & 75.0 & 54.7 & 76.2 & 11.8 & \underline{65.7} & \underline{48.9} & 71.9 & \underline{14.6} & 57.5 & 2.0 & 50.9 & 69.2 & 57.8 \\
\cA COLIPRI & 82.2 & 74.8 & 65.5 & 70.8 & 73.4 & 62.7 & 77.8 & 10.5 & 51.0 & 31.3 & 75.1 & 13.4 & 67.0 & 3.2 & 49.5 & \underline{70.1} & 64.3 \\
\cA Jolia & 75.3 & 66.6 & 71.3 & 76.2 & 62.0 & 51.4 & 72.8 & 9.4 & 61.4 & 36.0 & 65.8 & 8.9 & 71.2 & 4.2 & 57.8 & 65.8 & 57.1 \\
\cA \quad long prompts & 81.0 & 70.9 & 79.1 & 81.7 & 72.0 & 51.4 & 76.2 & 11.3 & 61.4 & 36.0 & 68.9 & 9.4 & 77.9 & 6.0 & 57.8 & 65.8 & 57.1 \\
\cA SPECTRE & 57.3 & 57.1 & 55.3 & 62.9 & 58.6 & 51.2$^{\P}$ & 55.2 & 11.3 & 50.4 & 32.8 & 55.0 & 4.8 & 54.2 & 0.0 & 50.9 & 60.5 & 69.4$^{\P}$ \\
\cA Merlin & 58.9 & 60.7 & 60.3 & 64.3 & 55.7 & 46.1 & 55.8 & 6.6 & 51.3 & 20.3 & 57.7 & 6.4 & 57.0 & 1.8 & 53.2 & 58.5 & 60.2 \\
\midrule
\cB fVLM & 77.7 & 67.7 & 55.3 & 54.8 & 67.0 & 52.3 & 70.2 & \underline{12.4} & 52.4 & 25.2 & 66.7 & 11.1 & 53.9 & $-$0.2 & 50.7 & 51.7 & 35.4 \\
\cB ViSD-Boost & 79.1 & 70.1 & 58.3 & 60.8 & 65.9 & 40.8 & 73.3 & 9.3 & 51.8 & 26.3 & 69.6 & 12.2 & 56.8 & 3.9 & 49.1 & 57.3 & 57.2 \\
\cB RADAR+, Merlin-only & 73.6 & 66.1 & 79.7 & 83.6 & 60.6 & 68.7 & 65.8 & 4.3 & 50.7 & 17.9 & 63.1 & 4.9 & 79.9 & \textbf{22.6} & 50.4 & 56.6 & 45.0 \\
\cB RADAR+, published & 70.3 & 64.7 & \underline{83.9} & \underline{88.2} & 59.4 & 68.9 & 63.4 & 2.4 & 52.9 & 22.3 & 60.5 & 6.4 & \underline{83.6} & 20.3 & \underline{57.8} & 62.5 & 58.5 \\
\midrule
\cC COLIPRI-FT & \underline{86.4} & \underline{77.4} & 82.7 & 85.0 & \underline{79.1} & \underline{72.9} & \underline{81.2} & 12.1 & 52.7 & 43.8 & \underline{76.5} & 10.4 & 81.6 & 5.8 & 52.0 & 68.4 & \underline{70.1} \\
\cD \textbf{ORGANIC-3D (ours)} & \textbf{87.7} & \textbf{79.7} & \textbf{87.7} & \textbf{90.1} & \textbf{79.5} & \textbf{75.0} & \textbf{82.5} & \textbf{28.5} & \textbf{76.9} & \textbf{74.7} & \textbf{78.2} & \textbf{18.4} & \textbf{86.5} & \underline{21.3} & \textbf{63.8} & \textbf{74.9} & \textbf{70.7} \\
\bottomrule
\end{tabular}
\\[1pt]
\parbox{\textwidth}{\scriptsize\setlength{\fboxsep}{1pt}%
Row colors: \colorbox{grpA}{no organ masks}, \colorbox{grpB}{organ masks at test time}, \colorbox{grpC}{fine-tuned on the same public data}, \colorbox{oursbg}{ours}. $^{\P}$AbdomenAtlas images are in SPECTRE's pretraining.}
\end{table*}
As shown in Fig. \ref{fig:overview}, ORGANIC-3D measures and introduces organ-level evidence dependence in 3D CT.
\cref{sec:tasks} defines four tasks to evaluate model performance and its image-text alignment capability, including classification, clinical finding independence, organ dependence, and laterality.
\cref{sec:encoder} presents the ORGANIC-3D encoder, which reads a 3D ViT through organ-specific anatomy queries and is trained to respond correctly to these interventions.
\cref{sec:vlm} presents the ORGANIC-3D VLM, which applies the same interventions to its answers.
\cref{sec:metrics} defines the evaluation metrics.

%-------------------------------------------------------------------------
\subsection{Image-text alignment tasks}
\label{sec:tasks}

\paragraph{Classification.}
\(V\) is a CT volume and \(y_f(V)\in\{0,1\}\) the report label of finding \(f\). An encoder scores \(f\) as
\begin{equation}
\Delta_f(V)=S\big(V,\text{“There is }f\text{.”}\big)-S\big(V,\text{“No }f\text{.”}\big),
\label{eq:delta}
\end{equation} 
where \(S\) is the image--sentence similarity. A VLM scores \(f\) by \(p_f(V)\), its probability of answering ``Yes''.

\paragraph{Clinical finding independence.}
A matched pair for a finding $f$ is two patients who differ in $f$ and agree on every other label,
\begin{equation}
  \begin{aligned}
  \Pi_f = \bigl\{(i,j) :{} & y_f(V_i) = 1,\ y_f(V_j) = 0, \\
  & y_{f'}(V_i) = y_{f'}(V_j)\ \forall f' \neq f \bigr\}.
  \end{aligned}
  \label{eq:pairs}
\end{equation}
Only a model that sees $f$ in the image scores the study with $f$ higher.

\paragraph{Organ dependence.}
To \emph{erase} organs \(U\), we fill their voxels dilated by 2 mm, with fat (\(-100\) HU) or soft tissue (\(40\) HU) plus noise: \begin{equation}
\mathcal{E}_{U}(V)=V\odot(1-M_U)+(c+\epsilon)\,M_U,\qquad \epsilon\sim\mathcal{N}(0,15^2).
\label{eq:erase}
\end{equation}

Erasing \(O(f)\) within the target organ removes the evidence for \(f\), while editing a control organ will not affect the evidence. Because any edit can shift a score on its own, we measure each effect as the change after erasing $O(f)$ minus the change after erasing a control organ. The 10 findings defined by size or shape (e.g.~cardiomegaly) are excluded. For lesion erasure evaluation on test sets with expert lesion masks, it replaces an annotated lesion with its organ's normal tissue and compares it with the same edit of a healthy region of the same organ.

\paragraph{Laterality.}
A sentence $s$ that names the side of a finding should score higher than the sentence with its side-swapped version, especially for lungs and pleura on chest and kidneys, bones and adnexa in the abdomen region. For example:

$s^{\leftrightarrow}$ (``\ldots in the left pleural space''):
  \begin{equation}
  S(V,s) > S\big(V,s^{\leftrightarrow}\big).
  \label{eq:side}
  \end{equation}

During the evaluation stage, \(V\) stays the same and only sentences are modified. We report laterality accuracy when evaluating the encoders.

%-------------------------------------------------------------------------
\subsection{ORGANIC-3D encoder}
\label{sec:encoder}

\paragraph{Architecture.}
The image backbone $E_\phi$ is the 3D vision transformer from the released COLIPRI~\cite{colipri} (Primus-M, 144M parameters).
A $192^3$ CT volume at 2\,mm generates $24^3$ patch tokens. We introduce 144 learned vectors as anatomy queries, 8 per organ group (16 groups; left and right lung and kidney separate) and 16 for global. Through two cross-attention blocks, each query extracts information from all patch tokens and generates one anatomy token $T_k$ for the scan.

\paragraph{Organ routing.}
Instead of pooling the tokens of all organs, we align each report sentence or finding prompt $s$, with the corresponding organ token, $\mathcal{K}(s)$:
\begin{equation}
  S(V,s)=\frac{1}{\tau}\cos\Big(g\Big(\sum_{k\in\mathcal{K}(s)}\beta_k(s)\,T_k\Big),\;h(t_s)\Big),
  \label{eq:score}
  \end{equation}
where $t_s$ is the CXR-BERT \cite{boecking2022biovil} embedding of $s$, \(\beta_k(s)\) are softmax weights over $\mathcal{K}(s)$ from each token's similarity to the sentence as in text-conditioned pooling \cite{mukhoti2023pacl}, \(g\) and \(h\) are projection heads, and \(\tau\) is a learned temperature. Unlike segmenter gating \cite{radar2026, shui2025fvlm}, no segmentation map is applied during organ routing.

\paragraph{Encoder training losses.} 
In addition to the contrastive loss between scans and report sentences, the encoder is trained with four ORGANIC losses. Each evaluates how a finding's score changes when a training scan is edited \cite{chang2021counterfactual, geiger2022iit}.

\emph{Organ erasure loss}. When the finding's corresponding organ is erased, the scan no longer shows whether $f$ is present. Its score should move to the base rate $b_f$ of $f$, initialized from the prevalence of $f$ in the training scans and learned. Meanwhile, erasing a control organ should leave the score unchanged. With $V^{\mathrm{t}}$ and $V^{\mathrm{c}}$ the scan with its target organ or a control organ erased:  
  \begin{equation}
    \mathcal{L}_{\mathrm{ev}}=\rho\big(\ell_f(V^{\mathrm{t}})-b_f\big)+\rho\big(\ell_f(V^{\mathrm{c}})-\operatorname{sg}[\ell_f(V)]\big)
    \label{eq:ev}
  \end{equation}
where $\ell_f=\Delta_f+c_f$ is the finding score plus a learned per-finding bias $c_f$ and $\rho$ is the Huber loss, and sg represents the stop gradient, so the original scan's score serves as a fixed reference for the control copy rather than moving toward it.

\emph{Mirror loss}. On a left–right flipped training scan, the side-swapped sentence must score higher than the original sentence.

\emph{Matched pair loss}. The patient with $f$ must score higher than its matched partner Eq.~\eqref{eq:pairs}.

\emph{Anatomy attention loss}. Each organ query extracts information from all patch tokens. This loss raises the total weight on the patch tokens inside its organ's mask. The mask is not applied to the attention itself, so the organ's token is built mainly from its organ.

 The mirror and matched pair losses are logistic ranking losses, $-\log\sigma(a-b)$. The anatomy attention loss is $-\log$ of the attention share inside the mask. The organ erasure loss starts after 20\% of training. No masks or edits enter the model at test time.

%-------------------------------------------------------------------------
\subsection{ORGANIC-3D VLM}
\label{sec:vlm}
% Table 2: VLM yes/no questions. Load with \input{sec/tables/tab_vlm}.
\begin{table*}[!t]
\centering
\footnotesize
\setlength{\tabcolsep}{3.2pt}
\renewcommand{\arraystretch}{0.9}
\setlength{\aboverulesep}{1.5pt}\setlength{\belowrulesep}{1.5pt}
\setlength{\abovecaptionskip}{2pt}
\definecolor{oursbg}{RGB}{226,234,248}%  ORGANIC-3D rows (same light blue as Fig. 2)
\definecolor{fail}{RGB}{250,222,218}%    at or below chance on an alignment test
\newcommand{\cD}{\rowcolor{oursbg}}%
\newcommand{\cE}{\rowcolor{oursbg!45}}%
\newcommand{\fl}[1]{\cellcolor{fail}#1}%
\caption{\textbf{VLM evaluation} (all values in \%). Visual question answering on the full CT-RATE (3,002) and Merlin (5,109) test sets and 464 ReXGroundingCT scans. Reported metrics include answer AUROC, BACC and F1 of the stated answers; MP and OEH as in Sec.~\ref{sec:metrics}. Bold: best; underlined: second best.}
\label{tab:vlm}
\begin{tabular}{@{}l ccccc ccccc ccccc@{}}
\toprule
 & \multicolumn{5}{c}{CT-RATE} & \multicolumn{5}{c}{Merlin} & \multicolumn{5}{c}{ReXGroundingCT} \\
\cmidrule(lr){2-6} \cmidrule(lr){7-11} \cmidrule(lr){12-16}
Model & AUROC & BACC & F1 & MP & OEH & AUROC & BACC & F1 & MP & OEH & AUROC & BACC & F1 & MP & OEH \\
\midrule
Hulu-Med & 80.8 & 63.5 & 38.5 & 73.1 & 59.8 & 70.0 & 55.2 & 12.4 & 69.6 & 10.0 & 72.9 & 60.4 & 34.5 & 66.8 & 66.9 \\
OmniCT & 80.5 & 68.1 & 45.5 & 73.5 & 57.7 & 66.9 & 56.2 & 19.1 & 66.6 & 10.2 & 71.2 & 62.9 & 39.8 & 66.3 & \underline{69.1} \\
MedGemma 1.5 & 71.5 & 60.0 & 34.2 & 64.7 & 28.6 & 66.9 & 57.2 & 20.3 & 66.0 & 17.9 & 59.6 & 55.0 & 26.2 & 58.0 & 28.9 \\
M3D-LaMed & 57.9 & 53.6 & 31.8 & 52.8 & 40.0 & 53.8 & 51.1 & 14.0 & 53.6 & -0.7 & 54.2 & 51.4 & 27.5 & 50.7 & 54.0 \\
RadFM & 54.2 & 50.4 & 21.4 & 52.4 & 18.6 & 53.7 & 50.2 & 7.9 & 56.7 & 4.9 & 53.9 & 50.4 & 19.9 & 48.6 & 17.2 \\
CT-CHAT & 67.5 & 53.2 & 33.9 & 58.8 & 11.0 & 53.4 & 49.9 & 17.1 & 52.6 & 1.1 & 58.8 & 53.5 & 35.0 & 56.1 & 10.6 \\
NV-Reason-CT & \underline{86.3} & \underline{78.0} & \textbf{60.2} & \underline{80.9} & \underline{67.4} & 74.1 & 67.8 & 27.0 & 73.8 & 46.4 & \underline{78.6} & \underline{70.4} & \underline{48.3} & \underline{71.7} & 65.2 \\
\midrule
Ours, plain fine-tuning & 82.1 & 72.3 & 52.2 & 76.1 & 30.7 & \underline{81.3} & \underline{68.2} & \underline{32.9} & \underline{80.5} & \underline{46.6} & 75.4 & 65.3 & 44.3 & 67.7 & 39.1 \\
\cD \textbf{ORGANIC-3D VLM (ours)} & \textbf{86.3} & \textbf{78.9} & \underline{58.7} & \textbf{82.8} & \textbf{96.8} & \textbf{87.6} & \textbf{80.0} & \textbf{40.6} & \textbf{86.0} & \textbf{96.2} & \textbf{79.5} & \textbf{72.1} & \textbf{49.0} & \textbf{74.1} & \textbf{98.9} \\
\bottomrule
\end{tabular}
\end{table*}

\paragraph{Architecture.} We use the trained ORGANIC-3D encoder as a frozen 3D vision encoder for VLM post-training. A two-layer MLP projects each of anatomy tokens $T_k$ into a visual token in the embedding space of the Qwen3.5-4B language model~\cite{qwen35}, instead of its 2D vision tower. Visual tokens are arranged in a fixed organ order to preserve organ identity through their positions. Only the MLP and the language model's LoRA adapter~\cite{hu2022lora} are trained.

\paragraph{Training.} The VLM training loss $\mathcal{L}_{\mathrm{LM}}$ is the next-token cross-entropy on the binary answers to the question $q_f$ whether $f$ is present, and on answer-first reports, which first list the labelled findings and then write the report text (``Abnormal findings: \ldots. Report: \ldots''). Because this loss can be met without reading the scan, we add an evidence-dependent loss that aligns each response to the image. With $a_i$ the correct answer for patient $i$ and $d(V')=\log p(V_i)-\log p(V')$ how much more likely that answer is on the patient's own scan $V_i$ than on another scan $V'$,

\begin{equation}
  \mathcal{L}_{\mathrm{dep}}=\sum_{V'\in\{V_j,\,V_i^{\mathrm{t}}\}}\max\big(0,\,m-d(V')\big) + \gamma\,\big|d(V_i^{\mathrm{c}})\big|,
  \label{eq:dep}
  \end{equation}

Once the finding's organ is erased for patients with $f$, the model should less likely to answer "Yes". Meanwhile, when a control organ is erased, the answer should stay unchanged and it penalizes answering "No" to any edited scan. An answer-attention loss rewards the model for directing more of its attention over the 144 anatomy tokens to the target organ's tokens, and a report margin requires each training report to be more likely paired to its own scan than another patient's scan.

%-------------------------------------------------------------------------
\subsection{Alignment metrics}
\label{sec:metrics}

All metrics are computed on test patients per finding, averaged over findings and reported in percentage. Higher scores indicate better performance. Here are the standard metrics, including multi label classification (MLC) AUROC of $\Delta_f$, zero-shot AUROC of tumor detection (liver, pancreas and kidney tumors), and lesion AUROC with labels from expert masks; for VLMs, answer AUROC of $p_f$, the balanced accuracy and F1; for report generation, F1 and CRG ~\cite{hamamci2025crg} of the extracted finding labels, and GREEN ~\cite{ostmeier2024green} and RadGraph F1 ~\cite{jain2021radgraph} against the reference report.

\emph{Alignment metrics}. The \emph{matched pair (MP) AUROC} is the AUROC over the matched pairs of Eq.~\eqref{eq:pairs}~\cite{janes2009aroc}. The \emph{organ erasure (OE) $\Delta$AUROC} is the AUROC lost after erasing the target organ minus that lost after erasing a control organ. For VLMs, the \emph{organ erasure hit (OEH) rate} is the same difference for the share of correctly stated findings that the model withdraws, and OE is computed on probability of "Yes". The \emph{lesion erasure (LE) hit rate}  is the share of expert-segmented lesions whose erasure lowers the lesion's sentence score more than the control edit. \emph{Laterality accuracy} (Side) is the share of side-naming sentences that score above their side-swapped version on the original scan, averaged over left and right sentences. The \emph{lung lobe erasure hit rate} (LL) is the share of single-lobe sentences whose score drops most when that lung lobe is erased. Image-attributable F1 (IA-F1) is the F1 of a report composed from the answers minus the F1 the same report gets on another patient~\cite{babar2021unconditioned}.

\section{Experiments}
\label{sec:experiments}

%-------------------------------------------------------------------------
\subsection{Datasets}
\label{sec:datasets}
We train on the public training splits of CT-RATE \cite{hamamci2024ctrate} (22,757 chest CT scans of 18,871 patients) and Merlin dataset \cite{blankemeier2024merlin} (15,314 abdominal CT studies). The released Merlin label file leaves 74\% of the positive findings unlabelled, so we re-extract test labels with evidence sentence from the full report using Qwen-3.8-27B and confirming with GPT-6-Luna. This resolves findings that the released labels miss when a report describes them in other words, such as an "enlarged liver" for hepatomegaly.

We test on the test splits of CT-RATE (3,002 volumes) and Merlin (5,109 studies) and on three other public datasets, including RAD-ChestCT \cite{draelos2021radchest} (3,630 chest volumes), and two datasets with radiologist-annotated lesion masks for lesion-level tests, ReXGroundingCT \cite{rexgroundingct} (chest lesions in 464 CT-RATE test scans) and AbdomenAtlas 3.0 \cite{bassi2025radgpt} (liver, pancreas and kidney tumors in 926 test scans).

%-------------------------------------------------------------------------
\subsection{Implementation details and baselines}
\label{sec:implementation}
\paragraph{Training.} 
\emph{Encoder:} The encoder is trained for 12,000 steps on four H100 GPUs with a batch size of 4 per GPU, which takes 22 hours. 3D CT studies are drawn from CT-RATE and Merlin in a ratio of 3 to 2 and cropped to $192^3$ voxels at 2\,mm. We use AdamW with a learning rate of $2\times10^{-5}$ for the pretrained backbone and $2\times10^{-4}$ for the new layers, with cosine decay. Every scan in each batch is used for the contrastive and classification losses. A quarter of all scans are selected as matched pairs (Eq. \ref{eq:pairs}) for the matched pair loss. Independently, three quarters of all scans receive a target-erased and a control-erased copy for the organ erasure loss, and half of the side-naming sentences are paired with a flipped copy of their scan for the mirror loss. These copies are built from the batch's scans and are not counted in the batch size.

\emph{VLM:} The VLM is trained for 10,000 steps on four B200 GPUs with a batch size of 4 per GPU, which takes 4.4 hours. It uses LoRA of rank 32 and a learning rate of $2\times10^{-4}$, and half of each batch are matched pairs. In Eq.~\eqref{eq:dep} we set $m=1$ and $\gamma=0.5$.  For both models we use the last checkpoint, without checkpoint selection. All settings are listed in supplement materials.

\paragraph{Inference.} The encoder scores each finding by Eq.~\eqref{eq:delta}. For VLM, the answer score $p_f$ is the probability of ``Yes'' against ``No'', and it claims a finding when $p_f\geq0.5$. For report generation, the model first writes the claimed findings. Without guidance, it often skips from this list to a numbered impression, so we start the report with anatomy prefix (``Lower thorax:'' on Merlin, ``Trachea'' on CT-RATE).

\paragraph{Baselines.} We compare eleven public vision language encoders with twelve checkpoints, including CT-CLIP and CT-VocabFine~\cite{hamamci2024ctrate}, fVLM~\cite{shui2025fvlm}, ViSD-Boost~\cite{visdboost}, HLIP~\cite{hlip}, ItemizedCLIP~\cite{itemizedclip}, COLIPRI~\cite{colipri}, Jolia~\cite{jolia2026}, SPECTRE~\cite{spectre}, Merlin~\cite{blankemeier2024merlin} and RADAR+~\cite{radar2026}. RADAR+ include two model weights, the Merlin-only checkpoint and the published checkpoint, pretrained on more than 400K in-house exams. We also fine-tune COLIPRI on CT-RATE and Merlin together as the controlled baseline. For generative VLM, we compare seven public models, including NV-Reason-CT-4B~\cite{nvreasonct2026}, which shares our language model and COLIPRI initialization but is trained on 550K instructions, OmniCT-7B~\cite{omnict}, Hulu-Med-7B~\cite{hulumed}, MedGemma 1.5-4B~\cite{medgemma15}, CT-CHAT-8B~\cite{hamamci2024ctrate}, M3D-LaMed-4B~\cite{bai2024m3d} and RadFM-13B~\cite{wu2023radfm}. Three encoders (fVLM, ViSD-Boost and RADAR+) read each organ only inside a segmentation mask computed at test time, so part of their OE performance comes from this design rather than from learning the whole image volume. RADAR+ returns no score when its segmenter loses the erased organ. All models run with their published preprocessing methods and are scored with their own published prompts.

%-------------------------------------------------------------------------
\subsection{ORGANIC-3D Encoders}\label{sec:enc_results}

% Tab.~\ref{tab:1} (\label{tab:1}) is in tables_body.tex

% Table 3: report generation, single column. Load with \input{sec/tables/tab_reports}.
\begin{table}[!t]
\centering
\scriptsize
\setlength{\tabcolsep}{3.4pt}
\renewcommand{\arraystretch}{0.9}
\setlength{\aboverulesep}{1.5pt}\setlength{\belowrulesep}{1.5pt}
\setlength{\abovecaptionskip}{2pt}
\definecolor{oursbg}{RGB}{226,234,248}%  ORGANIC-3D rows (same light blue as Fig. 2)
\definecolor{fail}{RGB}{250,222,218}%    at or below chance on an alignment test
\newcommand{\cD}{\rowcolor{oursbg}}%
\newcommand{\cE}{\rowcolor{oursbg!45}}%
\newcommand{\fl}[1]{\cellcolor{fail}#1}%
\caption{\textbf{Report generation} (all values in \%) on the CT-RATE and Merlin test sets. Free text report: finding accuracy (F1, CRG) and similarity to the reference (GREEN~\cite{ostmeier2024green}, RadGraph F1). Finding-based report: F1 and image-attributable F1 (IA-F1, Sec.~\ref{sec:metrics}). Bold: best; underlined: second best.}
\label{tab:reports}
\begin{tabular}{@{}l cccc cc@{}}
\toprule
 & \multicolumn{4}{c}{Free text report} & \multicolumn{2}{c}{Finding-based report} \\
\cmidrule(lr){2-5} \cmidrule(lr){6-7}
 & \multicolumn{2}{c}{Findings} & \multicolumn{2}{c}{Similarity} & \multicolumn{2}{c}{Findings}  \\
\cmidrule(lr){2-3} \cmidrule(lr){4-5}\cmidrule(lr){6-7}
Model & F1 & CRG & GREEN & RadGraph & F1 & IA-F1 \\
\midrule
\multicolumn{7}{c}{\textit{CT-RATE}} \\
Hulu-Med & 31.2 & 39.4 & 36.6 & 14.0 & 53.3 & 30.0 \\
OmniCT & 38.4 & 41.5 & \textbf{46.9} & \textbf{25.6} & 52.5 & 28.8 \\
MedGemma 1.5 & 19.9 & 36.5 & 15.8 & 10.8 & 44.1 & 18.3 \\
M3D-LaMed & 11.6 & 34.3 & 1.7 & 2.2 & 34.2 & 5.0 \\
RadFM & 7.0 & 33.6 & 1.1 & 0.7 & 31.7 & 1.8 \\
CT-CHAT & 17.3 & 37.0 & \textbf{46.9} & \underline{24.9} & 40.5 & 13.0 \\
NV-Reason-CT & \underline{56.5} & \underline{49.6} & 40.6 & 18.1 & \textbf{60.0} & \underline{37.3} \\
\cD \textbf{ORGANIC-3D (ours)} & \textbf{56.8} & \textbf{52.9} & \underline{41.9} & 21.5 & \underline{59.6} & \textbf{37.6} \\
\midrule
\multicolumn{7}{c}{\textit{Merlin}} \\
Hulu-Med & 4.4 & 33.6 & 21.5 & 8.7 & 26.6 & 13.4 \\
OmniCT & 4.2 & 33.8 & 5.1 & 4.8 & 23.4 & 9.2 \\
MedGemma 1.5 & 9.1 & 35.4 & \underline{23.3} & 11.0 & 23.6 & 9.5 \\
M3D-LaMed & 3.6 & 33.7 & 1.4 & 3.4 & 17.8 & 1.6 \\
RadFM & 0.8 & 33.4 & 0.3 & 0.4 & 17.5 & 1.2 \\
CT-CHAT & 1.1 & 33.6 & 3.4 & 3.4 & 17.5 & 0.9 \\
NV-Reason-CT & \underline{26.2} & \underline{38.9} & 19.2 & \underline{19.1} & \underline{32.0} & \underline{20.2} \\
\cD \textbf{ORGANIC-3D (ours)} & \textbf{37.7} & \textbf{49.4} & \textbf{28.9} & \textbf{31.8} & \textbf{48.2} & \textbf{37.4} \\
\bottomrule
\end{tabular}
\end{table}

\paragraph{State-of-the-art classification compared to current benchmarks.} ORGANIC-3D achieves the highest MLC-AUROC on all three multi-label classification test sets, including the heldout RAD-ChestCT (Tab.~\ref{tab:encoders}). ORGANIC-3D is developed with public training data only, whereas the strongest baseline, the published RADAR+, is pretrained on more than 400K in-house exams. Our controlled baseline, COLIPRI fine-tuned on the same public data, underperforms ORGANIC-3D on every test set, Table \ref{tab:ablations} shows that this gain comes from the anatomy queries and organ routing, while the ORGANIC losses add organ dependence at no cost in accuracy.

\paragraph{Improved alignment transfers to tumors and lesions.} On AbdomenAtlas 3.0, which is an unseen held-out test set to most models except SPECTRE, ORGANIC-3D detects liver, pancreas and kidney tumors better than all released encoders. On the radiologist-segmented RexGroundingCT lesion, ORGANIC-3D also outperforms all public encorders.

\paragraph{Accuracy does not imply alignment.} Despite high accuracy, public encoders fail the alignment tests (Tab.~\ref{tab:encoders}). COLIPRI, the second-best public encoder on CT-RATE, can only detect the laterality of a finding within a scan by chance. ORGANIC-3D is the only encoder that is strong on all three tests. Its predictions rely least on co-occurring findings, drop most among ungated encoders when the finding's target organ is erased, and pick the correct side for side-naming sentences.

\paragraph{Learned alignment matches gating without its costs.} Segmenter-gated encoders obtain organ dependence through their design, reading each CT volume only inside the segmented organ. However, gating limits laterality understanding, such as RADAR+ assigns left and right kidneys to one class. ORGANIC-3D thus reaches the organ dependence of gating without masks at test time, and with the same training data it is also more accurate.

\paragraph{Improved lesion-level alignment.} Lesion erasure is an edit that our model is not trained for. When an expert-segmented lesion is replaced by its organ’s normal tissue, ORGANIC-3D still has a higher lesion erasure hit rate score than the public encoders on two lesion erasure test sets, RexGroundingCT and AbdomenAtlas 3.0.

\subsection{VLMs}\label{sec:vlm_results}

% Tab.~\ref{tab:2} (\label{tab:2}) is in tables_body.tex

\paragraph{Accurate VQA with far less training data.} Compared to 7 public 3D VLMs, ORGANIC-3D VLM is level with the strongest public VLM, NV-Reason-CT, on CT-RATE (AUROC 86.3 each), although NV-Reason-CT shares our language model and is trained on about 550K instructions from 70,111 CT scans of various resources. On Merlin, which is external for NV-Reason-CT, and on the radiologist-annotated ReXGroundingCT scans, ORGANIC-3D VLM leads every public VLM on every metric. 

\paragraph{Improved VLM image-text alignment.} When the organ of a correctly stated finding is erased, the ORGANIC-3D VLM withdraws 96–99\% of these findings beyond a control organ, compared to at most 69\% for the public VLMs. Also, its matched pair AUROC is higher than that of every public VLM in all three datasets. The same language model fine-tuned on the same encoder without our losses withdraws only 31–47\%, so this dependence comes from the ORGNIC VLM training loss design, not from the encoder alone.

\paragraph{Higher report generation accuracy.} We score two kinds of VLM generated report against the original radiologist's report (Tab.~\ref{tab:reports}). The free-text report is what the model directly writes, so we evaluate in two aspects, how accurate the model detects the findings and how similar its text is to the radiologist's report.We also build a finding-based report from each model's yes/no answers. On chest CT-RATE, ORGANIC-3D has higher performance than other public models (except NV-Reason-CT) in accuracy but underperforms in the similarity scores. On Merlin, our model leads in every metrics compared to other public models. 

\paragraph{Similarity metrics reward the reference style.} On CT-RATE, OmniCT and CT-CHAT write the reports most similar to the reference despite an F1 of only 38.4 and 17.3, because both are trained to reproduce CT-RATE's templated report text. On Merlin, ORGANIC-3D writes the most similar reports using anatomy prefix while not sacrificing the accuracy. 

\begin{figure*}[t]
\centering
\includegraphics[width=\linewidth]{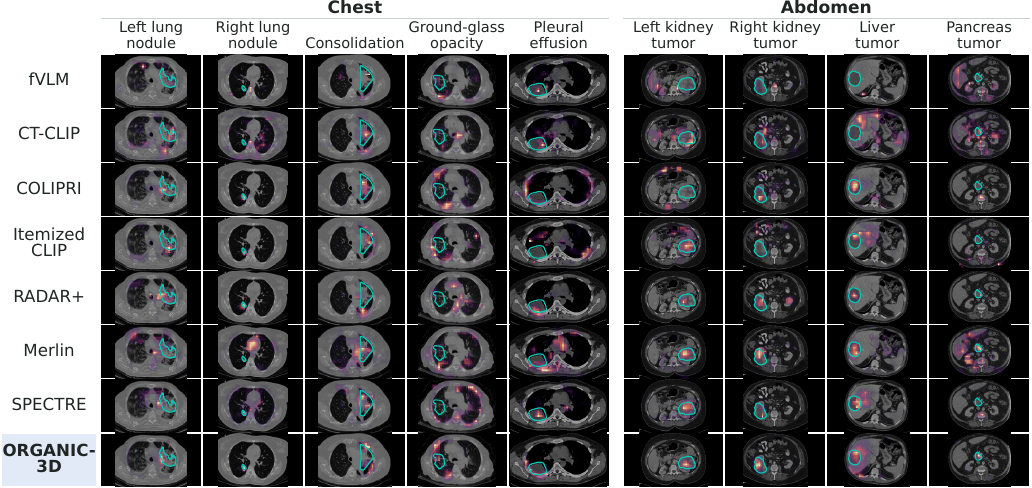}
\caption{\textbf{Evidence maps} (occlusion sensitivity) for nine findings (columns) and ORGANIC-3D and public encoders, one per family (rows; all checkpoints in the supplement). Brighter: removing the region lowers the finding's score more, and each map is scaled to its own maximum. Cyan: annotated lesion. Cases from ReXGroundingCT and AbdomenAtlas 3.0.}
\label{fig:3}
\end{figure*}

\subsection{Ablation studies}\label{sec:ablations}

% Tab.~\ref{tab:4} (\label{tab:4}) is in tables_body.tex

% Table 4: ablations, single column. Load with \input{sec/tables/tab_ablations}.
\begin{table}[!t]
\centering
\scriptsize
\setlength{\tabcolsep}{2.2pt}
\renewcommand{\arraystretch}{0.9}
\setlength{\aboverulesep}{1.5pt}\setlength{\belowrulesep}{1.5pt}
\setlength{\abovecaptionskip}{2pt}
\definecolor{oursbg}{RGB}{226,234,248}%  ORGANIC-3D rows (same light blue as Fig. 2)
\definecolor{fail}{RGB}{250,222,218}%    at or below chance on an alignment test
\newcommand{\cD}{\rowcolor{oursbg}}%
\newcommand{\cE}{\rowcolor{oursbg!45}}%
\newcommand{\fl}[1]{\cellcolor{fail}#1}%
\caption{\textbf{Ablation studies} (all values in \%). We compare COLIPRI and plain fine-tuning in encoder and VLM training to the anatomical-aware study design. OE indicates organ erasure $\Delta$AUROC, computed on the finding score $\Delta_f$ for the encoder and on P(Yes) for the VLM. Bold: best; underlined: second best.}
\label{tab:ablations}
\begin{tabular}{@{}l ccccccccc@{}}
\toprule
 & \multicolumn{3}{c}{MLC-AUROC} & Lesion & Tumor & \multicolumn{2}{c}{OE} & \multicolumn{2}{c}{Side} \\
\cmidrule(lr){2-4} \cmidrule(lr){5-5} \cmidrule(lr){6-6} \cmidrule(lr){7-8} \cmidrule(lr){9-10}
Encoder & CTR & MER & RAD & ReX & AA3 & CTR & MER & CTR & MER \\
\midrule
COLIPRI & 82.2 & 65.5 & 74.8 & 73.4 & 62.7 & 10.5 & 3.2 & 51.0 & 49.5 \\
+ fine-tuning & 86.4 & 82.7 & 77.4 & 79.1 & 72.9 & 12.1 & 5.8 & 52.7 & 52.0 \\
+ anat.\ queries & 87.1 & 85.5 & 79.4 & \textbf{79.9} & 74.4 & \underline{14.0} & 8.0 & 68.7 & \textbf{66.2} \\
+ organ routing & \underline{87.5} & \textbf{87.8} & \underline{79.6} & 79.0 & \textbf{75.6} & 13.5 & \underline{9.4} & \underline{74.9} & \underline{63.8} \\
\cD + ORGANIC (final) & \textbf{87.7} & \underline{87.7} & \textbf{79.7} & \underline{79.5} & \underline{75.0} & \textbf{28.5} & \textbf{21.3} & \textbf{76.9} & \underline{63.8} \\
\bottomrule
\end{tabular}
\\[3pt]
\begin{tabular}{@{}l cccccccc@{}}
\toprule
 & \multicolumn{2}{c}{Ans.\ AUROC} & \multicolumn{2}{c}{Report F1} & \multicolumn{2}{c}{MP} & \multicolumn{2}{c}{OE} \\
\cmidrule(lr){2-3} \cmidrule(lr){4-5} \cmidrule(lr){6-7} \cmidrule(lr){8-9}
VLM & CTR & MER & CTR & MER & CTR & MER & CTR & MER \\
\midrule
Plain fine-tuning & 82.1 & 81.3 & 22.6 & 2.1 & 76.1 & 80.5 & 9.0 & 9.4 \\
+ attention loss & 85.8 & 86.8 & \underline{27.4} & \underline{12.1} & 80.4 & 85.3 & 15.5 & 15.1 \\
+ dependence loss & \textbf{86.4} & \underline{87.3} & 19.8 & 10.9 & \underline{82.1} & \textbf{86.5} & \underline{21.4} & \textbf{23.6} \\
\cD + answer-first (final) & \underline{86.3} & \textbf{87.6} & \textbf{56.8} & \textbf{37.7} & \textbf{82.8} & \underline{86.0} & \textbf{28.1} & \underline{22.2} \\
\bottomrule
\end{tabular}
\end{table}

We ablate four distinct design axes of the ORGANIC-3D encoder and VLM respectively, shown in Tab.~\ref{tab:ablations}. For encoder development, the ablation studies test (i) fine-tuning on public dataset, (ii) anatomy query module, (iii) organ routing, and (iv) ORGANIC losses during training. For VLM development, it include (i) answer attention loss, (ii) evidence dependence loss, and (iii) answer-first report generation.
\paragraph{Anatomical structure adds accuracy and laterality.} Anatomy queries and organ routing raise accuracy further, most on Merlin and on the external RAD-ChestCT, and lift laterality accuracy to 74.9 on CT-RATE and 63.8 on Merlin, while the organ erasure $\Delta$AUROC stays low (13.5 CT-RATE and 9.4 Merlin).

\paragraph{The ORGANIC losses train alignment at no cost in accuracy.} Adding the ORGANIC losses doubles the organ-erasure $\Delta$AUROC, to 28.5 on CT-RATE and 21.3 on Merlin. Meanwhile, the classification, lesion and tumor detection, and laterality accuracy stay similar performance. The model also detects organ erasure stretegy it never saw in training. When the organ is filled with water or with its own median density instead of the trained fill, both the encoder and the VLM still respond far more strongly than models trained without the ORGANIC losses and NV-Reason-CT (supplement).

\paragraph{Each post-training component broadens VLM capability.} The answer attention loss makes answers more accurate with answer AUROC from 82.1 on CT-RATE and 81.3 on Merlin to 85.8 and 86.8. The evidence dependence loss makes them depend on the image with higher matched pair AUROC and organ erasure $\Delta$AUROC. The answer-first report training improve free-text report F1 without reducing other performance.

\subsection{Qualitative analysis}\label{sec:qualitative}
Fig.~\ref{fig:3} shows nine findings that cover the major organs of chest and abdominal CT and the key finding of each region, including lung nodules, consolidation, ground-glass opacity, pleural effusion, and kidney, liver and pancreas tumors. To shows which image regions each encoder's score depends on, we compute occlusion sensitivity maps~\cite{zeiler2014visualizing} on cases from RexGroundingCT and AbdomenAtlas 3.0 which have radiologist-annotated lesion maps. A 24\,$\times$\,24\,mm block of normal tissue (median lung density for lung region, soft tissue within organ) slides across the slice through the lesion in 16\,mm steps, and each position shows how much the score of the finding's sentence drops. Occlusion sensitivity maps only read each model's output score and each map is normalized to show where the encoders look at the scan. For ORGANIC-3D, the evidence lies within or on the edge of the annotated lesion for both the chest and abdomen findings, and spreads over the affected lung for ground-glass opacity. Public encoders often depend on regions away from the lesion, such as the mediastinum for a right lung nodule (Merlin) or the liver for a pancreas tumour (fVLM). Gated encoders read each organ inside a segmentation mask, so their maps are partly enforced to the organ by design. However, when RADAR+ segments both kidneys, it detects the right kidney tumor while highlighting the left kidney. Thus, the occlusion maps show image–text alignment at the lesion level that ORGANIC-3D makes a finding prediction depending on the region of the annotated lesion, while public encoders sometimes rely on regions away from it.

\section{Conclusion}\label{sec:conclusion}

Using matched patients, organ and lesion erasure, and side-swapped sentences, we investigated whether predictions from 3D CT VLMs rely on image evidence for each finding. Across eleven public encoders with twelve checkpoints and seven public VLMs, high diagnostic accuracy did not ensure that predictions depended on the corresponding organ or correct side. ORGANIC-3D explicitly trains for these properties through anatomy queries, organ routing, and the ORGANIC losses. Trained only on public data, its encoder achieves the highest classification AUROC among public encoders on every test set, including external evaluation on RAD-ChestCT and zero-shot tumor detection on AbdomenAtlas 3.0. It also achieves the strongest image–text alignment on every test except Merlin organ erasure, where the mask-gated RADAR+ scores slightly higher. Its VLM matches the strongest public CT VLM in answer accuracy on CT-RATE and surpasses it on Merlin and ReXGroundingCT. It withdraws 96--99\% of correctly stated findings after the corresponding organ is erased, compared with at most 69\% for public VLMs, and and leads in report finding accuracy. These results show that evidence dependence can be measured and learned without organ masks at test time, while improving diagnostic accuracy. ORGANIC-3D currently is limited to a fixed set of organ groups and findings, and only learns from report-derived labels. Extending it to lesion-level evidence and open vocabularies remains future work. We will release the benchmark, code, and model weights.
\section{LLM Usage Statement}
\label{sec:llm_usage}
Following the CVPR 2027 LLM policy, we disclose that LLM tools were used at the \emph{Assist} levels across writing, related work, code, experiments, figures, data, and research ideas. The authors reviewed, edited, and verified all LLM-assisted content, checked every citation, and take full responsibility for this paper.
{
    \small
    \bibliographystyle{ieeenat_fullname}
    \bibliography{main}
}


\end{document}